\documentclass{article}
\usepackage{ijcai24}

\usepackage{times}
\usepackage{soul}
\usepackage{url}
\usepackage[hidelinks]{hyperref}
\usepackage[utf8]{inputenc}
\usepackage[small]{caption}
\usepackage{graphicx}
\usepackage{amsmath}
\usepackage{amsthm}
\usepackage{booktabs}
\usepackage{algorithm}
\usepackage{algorithmic}
\usepackage[switch]{lineno}

\usepackage{amssymb}

\usepackage{cleveref}
\usepackage{biblatex}
\bibliography{ijcai24}

\usepackage{subcaption}
\usepackage{tabularx}
\usepackage{afterpage}
\usepackage{threeparttable}
\usepackage{sidecap}
\usepackage{colortbl}
\usepackage{xcolor}
\usepackage{multirow}
\definecolor{mycolor}{RGB}{219,219,219} 
\usepackage{diagbox}
\definecolor{yourcolor}{RGB}{220,224,228}
\definecolor{hiscolor}{RGB}{220,224,228}

\usepackage{multirow}
\usepackage{graphicx} 
\usepackage{float} 
\usepackage{epstopdf}
\usepackage{epsfig}

\title{APLe: Token-Wise Adaptive for Multi-Modal Prompt Learning}

\author{
Guiming Cao$^1$
\and
Kaize Shi$^1$\and
Hong Fu$^{2}$\and
Huaiwen Zhang$^3$\And
Guandong Xu$^{1,2}$\\
\affiliations
$^1$University of Technology Sydney\\
$^2$The Education University of Hong Kong\\
$^3$Inner Mongolia University\\
}

\begin{document}

\maketitle

\begin{abstract}
Pre-trained vision-language (V-L) models set the benchmark for generalization to downstream tasks among the noteworthy contenders. Many characteristics of the V-L model have been explored in existing research including the challenge of the sensitivity to text input and the tuning process across multi-modal prompts. With the advanced utilization of the V-L model like CLIP, recent approaches deploy learnable prompts instead of hand-craft prompts to boost the generalization performance and address the aforementioned challenges. Inspired by layer-wise training, which is wildly used in image fusion, we note that using a sequential training process to adapt different modalities branches of CLIP efficiently facilitates the improvement of generalization. In the context of addressing the multi-modal prompting challenge, we propose \textit{token-wise \textbf{A}daptive for multi-modal \textbf{P}rompt \textbf{L}earning} (APLe) for tuning both modalities prompts, vision and language, as tokens in a sequential manner. APLe addresses the challenges in V-L models to promote prompt learning across both modalities with adaptation, which indicates a competitive generalization performance in line with state-of-the-art baselines. Preeminently, APLe shows robustness results in prompt length and overfitting experiments with an advantage by adopting CLIP zero-shot knowledge to token-wise knowledge training framework.
\end{abstract}

\begin{figure}[htbp]
  \centering
    \includegraphics[width=0.9\linewidth]{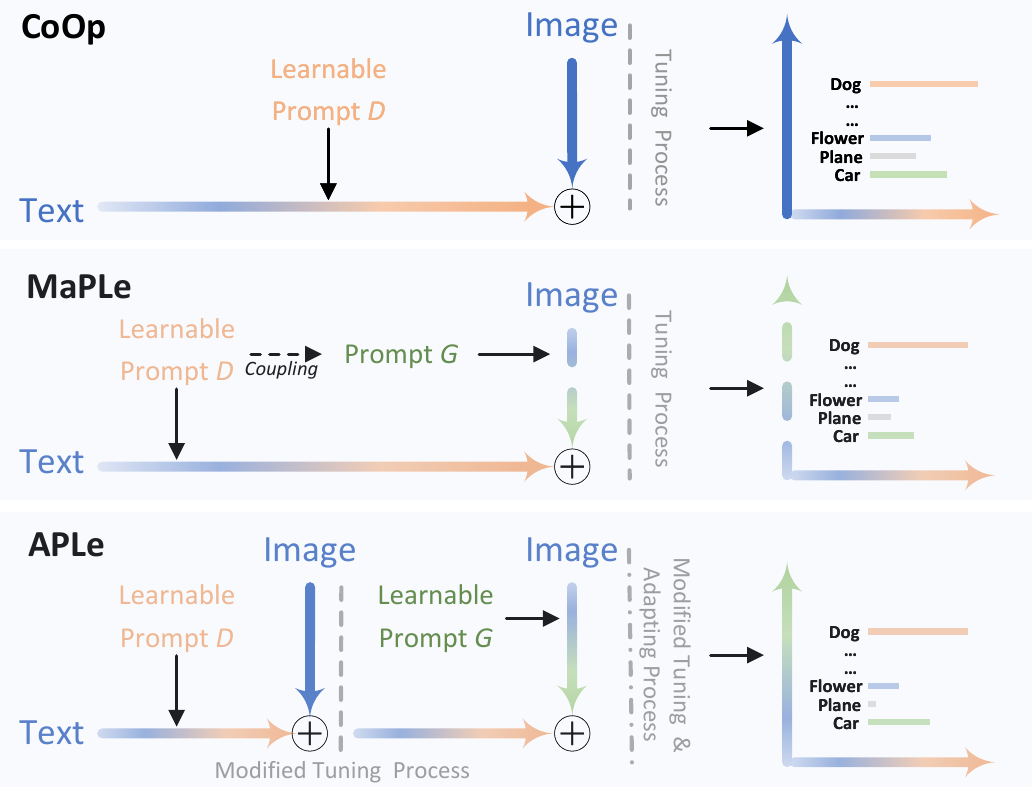} 
\captionsetup{labelsep=period,skip=2pt}
\captionsetup{justification=raggedright,singlelinecheck=false}
\captionof{figure}{Comparison of framework. CoOp adopts a uni-modal prompting. MaPLe demonstrates prompt learning in a multi-modal manner by the coupling function. APLe proposes an independent and sequential multi-modal prompt learning with adaptation.}\label{fig1}
\vspace{-0.25cm} 
\end{figure}
\section{Introduction}
Vision-language (V-L) pretraining is an emerging paradigm that performs computer vision tasks, such as image recognition ~\cite{recognition1,recognition2,recognition3}, object detection~\cite{object-detection1,object-detection2,object-detection3,object-detection4}, and segmentation~\cite{segmentation1,segmentation2,segmentation4}, while it inspired general application in other research \cite{9923419,shi2023amr,shi2023llama}, inspirational elevating the performance in generalization ability. A novel model, CLIP~\cite{CLIP} indicates the success of the paradigm by leveraging contrastive learning to align large-scale text-image pairs. This model introduces a simple inference process that pairs text embedding from a hand-crafted query and image embedding by calculating cosine similarity to predict the class of the image, which demonstrates the excelling generalization capability in the tasks. 

CLIP triggers the generalization toward a new era with potential. An earlier study, CoOp~\cite{COOP}, introduced a learnable continuous prompt approach as a text input instead of hand-crafted prompts shown in Fig.~\ref{fig1}. CoOp further reveals the potential and challenge of the vision-language pretraining paradigm by delivering exceptional results on the accuracy of image classification, while it also tackles the challenges of the extensive time demanded for training. Recent-coming research, MaPLe~\cite{multi-modal}, argues that it is the insufficient generalization performance for the downstream task if only the language prompt is explored. Thus, MaPLe exploits the prompting learning in both modalities shown in Fig.~\ref{fig1}, while MaPLe initiates vision prompts by transforming language prompts in a coupling function.

Despite the effectiveness of MaPLe toward sufficiently exploiting both behaviors of both modalities to contrastive learning. Its coupling mechanism makes it infeasible to conduct prompt learning in two modalities equally due to the disparate sophistication and conditions. Such mechanisms potentially introduce challenging problems resulting in generalization capability, such as overfitting caused by the prompt length. To this end, inspired by layer-wise training and image fusion that contributes to separately extracting and processing the different features from images. This paper proposes token-wise \textbf{A}daptive for multi-modal \textbf{P}rompt \textbf{L}earning (APLe) to adopt multi-modal prompt learning independently and sequentially to tackle the challenges and boost generation capability with the framework demonstrated in Fig.~\ref{fig1}. 

In the fundamental experiments for generalization evaluation, APLe demonstrated a comparable performance with the existing methods, particularly in datasets with domain shifts. Distinctively, APLe exploits the knowledge from both modalities and illustrates the advantage of robustness by sequential prompt learning and image adaptation, while the existing approaches are insufficient or omit the image conditions and multi-modal prompt conditions. 
Thanks to its effectiveness, APLe attains competitive generalization capability over the baselines by effectively addressing the challenges caused by image complexity and prompt length. To sum up, the main contributions of the paper are summarized as follows: 
\begin{itemize}
    \item We propose the token-wise adaptive for multi-modal prompt learning to adopt the V-L model. To the best of our knowledge, this is the first approach to address the overfitting caused by the prompt length and the disparate sophistication of features caused by multiple modalities.
    \item To extend the generalization capability, The paper proposes the token-wise knowledge training framework that allows a prompt learning process to be incorporated with CLIP zero-shot knowledge in a prompt-wise manner. Such a framework promotes APLe to learn the knowledge from two modalities without interference, which attains adaptive and stable generalization capability.
    \item This approach conducts prompt learning with zero-shot knowledge followed by a token adaptation function to alleviate knowledge conflicts and encourage collaboration between modalities, thereby the adaptation function encourages APLe to align the V-L representation.
    
\end{itemize}    
\begin{figure*}[t] 
\centering 
\includegraphics[width=0.78\textwidth]{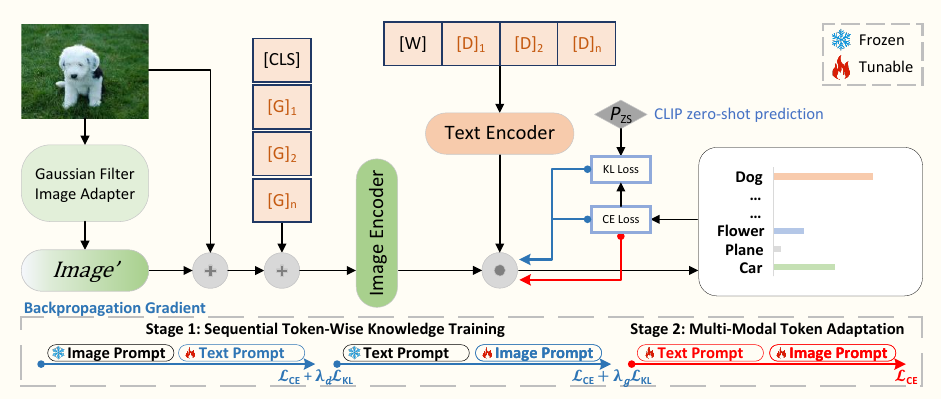} 
\captionsetup{labelsep=period,skip=2pt}
\captionsetup{justification=raggedright,singlelinecheck=false}
\captionof{figure}{Overview of APLe (Token-Wise Adaptive for Multi-Modal Prompt Learning) framework for prompting. APLe first deploys an image adapter to mitigate image noise and enhance image features effectively. Then, the prompts were trained sequentially as tokens with CLIP zero-shot knowledge and adaptation to alleviate the knowledge conflicts and prompt the synergy between modalities. Specifically, CE denotes the cross-entropy loss, KL denotes the Kullback-Leibler loss and $\lambda$ presents the hyper-parameter used to combine the loss functions. }\label{Fig.main1} 
\vspace{-0.25cm}
\end{figure*}

\section{Related Work}
\textbf{Vision Language Models.} The multi-modalities research in exploring vision-language interaction is a core topic. At the early stage, a few attention-based approaches are introduced including BAN~\cite{BAN}, Intra-Inter~\cite{Intra-Inter}, and MCAN~\cite{MCAN}, which make contributions to the downstream tasks. While the following approaches are proposed with BERT-like architecture, ViLBERT~\cite{ViLBERT}, LXMERT~\cite{LXMERT} and UNITER~\cite{UNITER}, expand the boundary in reasoning between the modalities. In the near past, the two prominent models, CLIP and ALIGN~\cite{ALIGN} have dominated the vision-language tasks through representation learning with an impressive generalization performance.

To adopt the V-L model in downstream tasks, the conventional approaches do not satisfy the current preference, which constrains the zero-shot or few-shot capability by fine-tuning and linear probing. Thanks to the supervised training with abundant data from the internet, $\sim$400M image-text pairs used in CLIP and $\sim$1B used in ALIGN, both models demonstrated the power of visual-language pre-training paradigm in zero-shot learning and few-shot learning. Inspired by CLIP and ALIGN, many recent proposed approaches achieve dominant generalization performance on specific tasks by adopting the V-L paradigm into downstream tasks including few-shot image recognition, object detection, and segmentation. 
\newline  
\textbf{Prompt Design.} The prompt design inspires research interest by the brand new paradigm, ``pre-train, prompt, and predict'', which was used in NLP tasks~\cite{NLP1,NLP2,NLP3} at the early stage. The recently proposed approaches introduced the prompt learning paradigm to adopt the V-L model to downstream tasks which takes a query, as known as a prompt, into the language branch to boost the generalization capability in downstream tasks. Therefore, regarding the effectiveness of prompt design, a rising trend of research lies on prompt design or prompt learning to further extend the potential of the V-L models and address the newly arisen challenges caused by prompt learning, such as the language-only prompting model CoOp, vision-only prompting model VPT ~\cite{vision1} and a recently proposed method, MaPLe, in multi-modal prompting. 
\newline
\textbf{Prompt Learning in Vision and Language.} The early work CoOp exploits a learnable token into a language branch for fine-tuning CLIP. Its subsequent work Co-CoOp~\cite{COCOOP} further extends the boundary of generalization by exploiting conditional prompt. For the works applying prompt learning in the vision branch, VPT is the first work to perform vision prompt in CLIP while~\cite{video18} introduces the prompt learning paradigm into video understanding tasks. The recent work, MaPLe, proposed a framework that applies learnable tokens into both vision and language branches by coupling function. MaPLe shows an advanced generalization performance owing to the coupling function improving the alignment between vision and language representations. 

\section{Method}
The proposed approach tackles the V-L pretraining model, CLIP, in downstream tasks. We argue that the two modalities, vision and language, equally contribute to downstream tasks while also delving into the challenges of disparate sophistication in prompt learning. 
To efficiently address the challenges, APLe proposes a framework that encourages prompts learned independently and sequentially from two modalities. 
Explicitly, APLe first attaches a Gaussian filter image adapter to alleviate the sophistication of the image features before prompt learning, which empowers the model to effectively mitigate image noise and enhance image features across various scales. 
Then, APLe is structured into two blocks. A sequential token-wise knowledge training block first explores learnable prompts as tokens for training with the CLIP zero-shot model, which obtains generalization knowledge and encourages multi-modality tokens to learn knowledge independently and sequentially without interference from the other. 
Ultimately, the multi-modal token adaptation block promotes the synergy between the tokens to adopt the V-L model.

\subsection{Revisiting CLIP}
\label{3.1}
 CLIP presents a basic architecture to adopt the vision-language model and addresses image recognition and classification tasks. Many existing approaches follow the architecture of CLIP. Many components used in CLIP including the neural network backbone, transformer and Vit (vision transformer), are inherited in the recent research and our approach, APLe. To adapt feature representation in CLIP, the text encoder transformed a hand-crafted query $Q$ with length $T$ as prompt into feature representations, and the image encoder progressed likewise to the image as features $I \in \mathbb{R}^{3 \times H\times W}$ for calculating cosine similarity and generating prediction.
\newline
\textbf{Text Feature Block.} Initially, the text query $Q$, ``a photo of \{class\}'', is tokenized and embedded. Then, the transformer layers $\mathcal{B}$ compute word embedding and forward its result to the projection function for the cosine similarity calculation.
\begin{equation}
\label{eq11}
    W = Embed(Tokenizer(Q))   \qquad W \in \mathbb{R}^{T\times d_{lang}}
\end{equation}
\begin{equation}
    W_n = \mathcal{B}_{n} (W_{n-1}) \qquad n = 1,2,3...,K.
\end{equation}
\begin{equation}
    Z_K = TextProj(W_K)  \qquad   Z\in \mathbb{R}^{d_{vis-lang}}
\end{equation}
\textbf{Image Feature Block.} Similar to text feature processing, the procedure excludes tokenization but appends a concatenation function for combining the class identification token $c$.
\begin{equation}
\label{eq1}
    E = PatchEmbed(I) \qquad 
\end{equation}
where E $\in \mathbb{R}^{M^2 \times {d_{vis}} }$ with the fixed-size patches M = 14 .
\begin{equation}
    \left [ c_n,E_n \right ] =\mathcal{V}_n([c_{n-1},E_{n-1}]) \qquad   n = 1,2,3...,K.
    \label{c_kkk}
\end{equation}
where $c$ denote as the token for class identification, $[\;,\;]$ as concatenation function, $\mathcal{V}$ as vision transformer layers.
\begin{equation}
    f = ImageProj(c_k)  \qquad   f\in \mathbb{R}^{d_{vis-lang}}
\end{equation}
Where $c_k$ is produced in $K$ steps in Equation.~\ref{c_kkk}. The result is transferred into the common V-L latent space by $ImageProj$ projection function for the final operation in Equation.~\ref{eq99}.\\
\textbf{Overall.} The text feature from prompts is transferred as $Z_{i }$, the image feature $f$ of the target image and the class to predict denoted as $i \in$ \{1,2...C\}, the final step of CLIP in cosine similarity calculation and predictions is shown below:
\begin{equation}
\label{eq99}
p_{i} =\frac{\mathrm{exp} (sim(Z_{i }\cdot f )/\tau )} { {\textstyle \sum_{j=1}^{C}} \mathrm{exp} (sim(Z_{j }\cdot f)) }
\end{equation}
where $\tau$ stands for the temperature parameter, $sim(\cdot)$ denoted as cosine similarity calculation and $p_{i}$ as the prediction result.
%

\subsection{APLe: Token-Wise Knowledge Training and Multi-Modal Token Adaptation}
\label{3.2}
This paper argues that the existing methods for uni- or multi-modalities carry out the learnable prompt training with weak considering the potential and challenge of the differences in vision and language modalities. To this end, we introduce token-wise training and adaptation framework to multi-modalities prompt learning which addresses the challenges by independently and sequentially training for both prompts and optimizing the utilization of the potential of the V-L model. While the holistic framework of APLe is shown in Fig.~\ref{Fig.main1}.\\ 
\textbf{Image Fusion with Adapter. }We note that disparate sophistication in images caused difficulty in prompt learning. Inspired by the existing work in image recognition tasks, a simple function is introduced to obtain a lucid and authentic image feature from the original image. The image $I$ in APLe is processed by Fast Fourier Transform $\mathcal{F}(\cdot)$ followed by the Gaussian filter $\mathcal{G}(\cdot)$ for suppressing image noise and enhancing subtle image features as the pipeline shown below:
\begin{align}
    \mathcal{G}(x,y)&=\frac{1}{2\pi\sigma^2} \cdot e^{-\frac{x^2+y^2}{2\sigma^2} }\\
  I^{'}&=  \mathcal{G}(\mathcal{F}(I))\label{eqgauss}
\end{align}%
Here $\sigma$ refers to the standard deviation in $\mathcal{G}(\cdot)$ and is set to 0.05 in our case, $x$ to the x-coordinate of each pixel in the image and $y$ to the y-coordinate of each pixel in the image.
\begin{align}
\label{eqibar}
  \bar{I}=\alpha I+(1-\alpha)\cdot \mathcal{F}^{-1}(I^{'} )
\end{align}%
Then, the image fusion operation is applied to the adjusted image and the original image where $\alpha$ is set to 0.9. 
The result in image fusion $\bar{I}$ is sequentially taken by image embedding. The fundamental aim of image fusion is to stabilize the variance of image features by suppressing interference rather than eliminating it. While image fusion alleviates the complexity of the image feature, the challenge is further efficiently compensated by vision tokens training in the APLe token-wise knowledge training and adaptation framework, which allows APLe to enhance its generalization performance further.
\begin{equation}
\label{eqfinal}
p_{i} =\frac{\mathrm{exp} (sim(Z_{i }\cdot \bar{f} )/\tau )} { {\textstyle \sum_{j=1}^{C}} \mathrm{exp} (sim(Z_{j }\cdot \bar{f})) }
\end{equation}
By introducing fusion image $\bar{I}$ into the image feature embedding pipeline, the new image feature $\bar{f}$ is taken into Eq.~\ref{eqfinal} for contrastive learning and predicting $p_i$ with text feature $Z$.\\
\textbf{Feature Embedding.} In contrast to the existing approach, APLe initiates learnable prompt $D$ and $G$ for text and image respectively, in a decoupling manner, which is embedded into their modalities feature by concatenation operation as seen in the Eq.~\ref{eqA2} and Eq.~\ref{eqsI2}. As the effectiveness of deep prompting, the component is maintained to APLe, which is deployed to process the newly concatenated prompts. In other settings, the length $m$ for both learnable tokens is set to 2 and the number of deep transformer layers is set with extra depth $S=8$, while $d_{lang} = 512$, $d_{vis} = 768$, and $d_{vis-lang} = 512$.\\
\textbf{Feature Embedding in Text Feature. }The encoding to text is similar to CLIP, but a learnable token $D$ is embedded.
\begin{equation}
\label{eqqqq}
    W = Embed(Tokenizer(Q))   \qquad W \in \mathbb{R}^{T\times d_{lang}}
\end{equation}
\begin{equation}
\label{eqA2}
\begin{split}
    &[W_s,D_s] = \mathcal{B}_{j} ([W_{s-1},D_{s-1}])  \\
&s = S+1,...,K. \qquad D \in \mathbb{R}^{m\times d_{lang}}
\end{split}   
\end{equation}
\begin{equation}
    Z_K = TextProj(W_K)  \qquad   Z\in \mathbb{R}^{d_{vis-lang}}
\end{equation}
\textbf{Feature Embedding in Image Feature. }The image encoding corresponds to CLIP, but the image is first processed by the image fusion followed by the embedding process. Then, the result is concatenated with the learnable vision token $G$.
\begin{equation}
\label{eqsI1}
    E = PatchEmbed(I) 
\end{equation}
\begin{equation}
\label{eqsI2}
\begin{split}
      &[ c_s,E_s,G_s  ] =\mathcal{V}_j([c_{s-1},E_{s-1},G_{s-1}]) \\
      &s = S+1,..,K. \qquad G \in \mathbb{R}^{m\times d_{vis}}
\end{split}
\end{equation}
\begin{equation}
\label{eqsI3}
  f = ImageProj(c_k)  \qquad   f\in \mathbb{R}^{d_{vis-lang}}
\end{equation}
In contrast with other approaches, the decoupling structure enables both vision and language prompts to learn knowledge separately, and alleviates the constraint from the other, which enhances the capability of multi-modal prompts learning.\\ 
\textbf{Training and Adaptation. } Preeminently, APLe explicitly and effectively boosts the generalization capability by conducting a distinctive and sequential training and adaptation process for prompt learning across multiple modalities. \\
\textbf{Sequential Token-Wise Knowledge Training. } APLe trains each modality prompt sequentially and separately as tokens at a time while keeping another one frozen. 
During the training, APLe incorporates CLIP zero-shot knowledge in the prompt learning to escalate generalization capability and mitigate the potential of overfitting due to prompt setting by using diverse combination parameters to language and vision prompts, $\lambda_d$ and $\lambda_{g}$ respectively, as demonstrated in Eqs.~\ref{stage1} -~\ref{stage11}.
\begin{align}
\label{stage1}
Stage1:\quad \mathcal{L}^{I}_{total}&=\mathcal{L}_{ce}(p_i,C_i) + \lambda_{d}\mathcal{L}_{kl}(p_i,p_i^{zs})\\
\label{stage11}
\mathcal{L}^{I}_{total}&=\mathcal{L}_{ce}(p_i,C_i) + \lambda_{g}\mathcal{L}_{kl}(p_i,p_i^{zs})
\end{align}
where $\mathcal{L}_{ce}$ and $\mathcal{L}_{kl}$ are the cross-entropy loss for classification prediction and Kullback-Leibler divergence loss for quantifying the prediction disparity between the trained model and CLIP zero-shot model, which are denoted as $p_i$ and $p_i^{zs}$ respectively. While separately training the modalities prompts, the hyper-parameter $\lambda$ is set for combining with KL loss, while  $\lambda_{d}$ and $\lambda_{g}$ are 0.5 for the language prompt training and set to 0.3 for the vision prompt training respectively.
\newline
\textbf{Multi-Modal Token Adaptation.} Owing to the token-wise knowledge training, APLe explores the potential of both modalities in prompt learning. For further adapting the synergy of the tokens from both branches and attaining advanced generalization capability in downstream tasks, we introduce a straightforward and transparent approach that retrains the model to promote synergy between the tokens without zero-shot knowledge as the end of the pipeline of APLe.
\begin{align}
Stage2:\quad \mathcal{L}^{II}_{total}&=\mathcal{L}_{ce}(p_i,C_i)
\vspace{-0.2cm} 
\end{align}
\section{Experiments}
To evaluate the capability of APLe, we deploy experiments with the aforementioned approaches, including CLIP, CoOp, Co-CoOp, and MaPLe. The experiment and the datasets used align with the baseline including three experiments and 15 datasets of Caltech101~\cite{Caltech101}, DTD~\cite{DTD}, EuroSAT~\cite{EuroSAT}, FGVCAircraft~\cite{FGVCAircraft}, Flowers102~\cite{Flowers102}, Food101~\cite{Food101},  ImageNet~\cite{ImageNet}, OxfordPets~\cite{OxfordPets}, StanfordCars~\cite{StanfordCars},  SUN397~\cite{SUN397}, and UCF101~\cite{UCF101}; ImageNetSketch~\cite{object-detection4}, ImageNetV2~\cite{ImageNetV2}, ImageNet-A~\cite{ImageNet-A} and ImageNet-R~\cite{ImageNet-R} with domain shift from the ImageNet dataset. 

\subsection{Fundamental Experiment }
To examine the zero-shot learning capability, the datasets are split into two groups, base classes and novel classes, with the evaluation of the average accuracy over 3 runs, and the harmonic mean (HM) of two accuracies is also included. \\
\textbf{Base-to-Novel Generalization.} APLe is trained on the base classes with 16-shot and then evaluated in both groups.\\
\textbf{Cross-Dataset Generalization.} For the evaluation of zero-shot learning and transfer learning, 
we train APLe in ImageNet by deploying all 1000 classes in a 16-shot manner. Then, the evaluation is conducted in another 10 datasets.\\
\textbf{Domain Generalization.} To further demonstrate the performance of APLe on generalization capability, 
The model trained in cross-dataset generalization is evaluated in the other four datasets with domain shift from the ImageNet.
\begin{table*}[t]
\vspace{-0.25cm}
\begin{minipage}[t]{0.3\textwidth}

\makeatletter\def\@captype{table}
\captionsetup{skip=2pt}
\caption*{\textbf{(a) Average over 11 datasets}}
\centering
\begin{tabular}{lccc}
    \toprule
    Name     & Base& Novel & HM\\
    \midrule
    CLIP& 69.34& 74.22& \multicolumn{1}{|c}{71.70}   \\
    CoOp& \textbf{82.69}& 63.22& \multicolumn{1}{|c}{71.66
} \\
    Co-CoOp& 80.47& 71.69& \multicolumn{1}{|c}{75.83
} \\  
    MaPLe& \underline{82.28}& \textbf{75.14}& \multicolumn{1}{|c}{\textbf{78.55}} \\   
    \midrule
    \rowcolor{mycolor}
    APLe& 81.99& \underline{75.11}& \multicolumn{1}{|c}{\underline{78.40}} \\  
    \bottomrule
\end{tabular}
\end{minipage}
\hspace{0.1in}
\begin{minipage}[t]{0.3\textwidth}
\makeatletter\def\@captype{table}
\captionsetup{skip=2pt}
\caption*{(b) ImageNet}
\centering
\begin{tabular}{lccc}
    \toprule
    Name     & Base& Novel & HM\\
    \midrule
    CLIP& 72.43& 68.14& \multicolumn{1}{|c}{70.22
}   \\
    CoOp& \underline{76.47}& 67.88& \multicolumn{1}{|c}{71.92
} \\
    Co-CoOp& 75.98& 70.43& \multicolumn{1}{|c}{73.10} \\  
    MaPLe& \textbf{76.66}& \underline{70.54}& \multicolumn{1}{|c}{\underline{73.47}} \\   
    \midrule
    \rowcolor{mycolor}
    APLe& 76.46& \textbf{70.78}& \multicolumn{1}{|c}{\textbf{73.51}} \\  
    \bottomrule
\end{tabular}
\end{minipage}
\hspace{0.1in}
\begin{minipage}[t]{0.3\textwidth}
\makeatletter\def\@captype{table}
\captionsetup{skip=2pt}
\caption*{(c) Caltech101}
\centering
\begin{tabular}{lccc}
    \toprule
    Name     & Base& Novel & HM\\
    \midrule
    CLIP& 96.84& 94.00& \multicolumn{1}{|c}{95.40}   \\
    CoOp& \underline{98.00}& 89.81& \multicolumn{1}{|c}{93.73
} \\
    Co-CoOp& 97.96& 93.81& \multicolumn{1}{|c}{95.84
} \\  
    MaPLe& 97.74& \underline{94.36}& \multicolumn{1}{|c}{\underline{96.02}} \\   
    \midrule 
    \rowcolor{mycolor}
    APLe& \textbf{98.30}& \textbf{94.80}& \multicolumn{1}{|c}{\textbf{96.52}} \\  
    \bottomrule
\end{tabular}
\end{minipage}
\newline

\begin{minipage}[t]{0.3\textwidth}
\makeatletter\def\@captype{table}
\captionsetup{skip=2pt}
\caption*{(d) OxfordPets}
\centering
\begin{tabular}{lccc}
    \toprule
    Name     & Base& Novel & HM\\
    \midrule
    CLIP& 91.17& 97.26& \multicolumn{1}{|c}{94.12
}   \\
    CoOp& 93.67& 95.29& \multicolumn{1}{|c}{94.47
} \\
    Co-CoOp& 95.20& \underline{97.69}& \multicolumn{1}{|c}{\underline{96.43}} \\  
    MaPLe& \textbf{95.43}& \textbf{97.76}& \multicolumn{1}{|c}{\textbf{96.58}} \\   
    \midrule
    \rowcolor{mycolor}
    APLe& \underline{95.22}& 97.56& \multicolumn{1}{|c}{96.37
} \\  
    \bottomrule
\end{tabular}
\end{minipage}
\hspace{0.1in}
\begin{minipage}[t]{0.3\textwidth}
\makeatletter\def\@captype{table}
\captionsetup{skip=2pt}
\caption*{(e) StanfordCars}
\centering
\begin{tabular}{lccc}
    \toprule
    Name     & Base& Novel & HM\\
    \midrule
    CLIP& 63.37& \textbf{74.89}& \multicolumn{1}{|c}{68.65
}   \\
    CoOp& \textbf{78.12}& 60.40& \multicolumn{1}{|c}{68.13
} \\
    Co-CoOp& 70.49& 73.59& \multicolumn{1}{|c}{72.01
} \\  
    MaPLe& \underline{72.94}& 74.00& \multicolumn{1}{|c}{\textbf{73.47}} \\   
    \midrule
    \rowcolor{mycolor}
    APLe& 72.42& \underline{74.42}& \multicolumn{1}{|c}{\underline{73.41}} \\  
    \bottomrule
\end{tabular}
\end{minipage}
\hspace{0.1in}
\begin{minipage}[t]{0.3\textwidth}
\makeatletter\def\@captype{table}
\captionsetup{skip=2pt}
\caption*{(f) Flowers102}
\centering
\begin{tabular}{lccc}
    \toprule
    Name     & Base& Novel & HM\\
    \midrule
    CLIP& 72.08& \textbf{77.80}& \multicolumn{1}{|c}{74.83
}   \\
    CoOp& \textbf{97.60}& 59.67& \multicolumn{1}{|c}{74.06
} \\
    Co-CoOp& 94.87& 71.75& \multicolumn{1}{|c}{81.71
} \\  
    MaPLe& \underline{95.92}& 72.46& \multicolumn{1}{|c}{\underline{82.56}} \\   
    \midrule
    \rowcolor{mycolor}
    APLe& 95.09& \underline{73.90}& \multicolumn{1}{|c}{\textbf{83.17}} \\  
    \bottomrule
\end{tabular}
\end{minipage}
\newline

\begin{minipage}[t]{0.3\textwidth}

\makeatletter\def\@captype{table}
\captionsetup{skip=2pt}
\caption*{(g) Food101}
\centering
\begin{tabular}{lccc}
    \toprule
    Name     & Base& Novel & HM\\
    \midrule
    CLIP& 90.10& 91.22& \multicolumn{1}{|c}{90.66
}   \\
    CoOp& 88.33& 82.26& \multicolumn{1}{|c}{85.19
} \\
    Co-CoOp& 90.70& 91.29& \multicolumn{1}{|c}{90.99
} \\  
    MaPLe& \underline{90.71}& \textbf{92.05}& \multicolumn{1}{|c}{\textbf{91.38}} \\   
    \midrule
    \rowcolor{mycolor}
    APLe& \textbf{90.74}& \underline{91.77}& \multicolumn{1}{|c}{\underline{91.25}} \\  
    \bottomrule
\end{tabular}
\end{minipage}
\hspace{0.1in}
\begin{minipage}[t]{0.3\textwidth}
\makeatletter\def\@captype{table}
\captionsetup{skip=2pt}
\caption*{(h) FGVCAircraft}
\centering
\begin{tabular}{lccc}
    \toprule
    Name     & Base& Novel & HM\\
    \midrule
    CLIP& 27.19& \textbf{36.29}& \multicolumn{1}{|c}{31.09
}   \\
    CoOp& \textbf{40.44}& 22.30& \multicolumn{1}{|c}{28.75
} \\
    Co-CoOp& 33.41& 23.71& \multicolumn{1}{|c}{27.74
} \\  
    MaPLe& \underline{37.44}& \underline{35.61}& \multicolumn{1}{|c}{\textbf{36.50}} \\   
    \midrule
    \rowcolor{mycolor}
    APLe& 36.65& 34.45& \multicolumn{1}{|c}{\underline{35.52}} \\  
    \bottomrule
\end{tabular}
\end{minipage}
\hspace{0.1in}
\begin{minipage}[t]{0.3\textwidth}
\makeatletter\def\@captype{table}
\captionsetup{skip=2pt}
\caption*{(i) SUN397}
\centering
\begin{tabular}{lccc}
    \toprule
    Name     & Base& Novel & HM\\
    \midrule
    CLIP& 69.36& 75.35& \multicolumn{1}{|c}{72.23
}   \\
    CoOp& \underline{80.60}& 65.89& \multicolumn{1}{|c}{72.51
} \\
    Co-CoOp& 79.74& 76.86& \multicolumn{1}{|c}{78.27
} \\  
    MaPLe& \textbf{80.82}& \textbf{78.70}& \multicolumn{1}{|c}{\textbf{79.75}} \\   
    \midrule
    \rowcolor{mycolor}
    APLe& 80.22& \underline{78.44}& \multicolumn{1}{|c}{\underline{79.32}} \\  
    \bottomrule
\end{tabular}
\end{minipage}
\newline

\begin{minipage}[t]{0.3\textwidth}
\makeatletter\def\@captype{table}
\captionsetup{skip=2pt}
\caption*{(j) DTD}
\centering
\begin{tabular}{lccc}
    \toprule
    Name     & Base& Novel & HM\\
    \midrule
    CLIP& 53.24& \textbf{59.90}& \multicolumn{1}{|c}{56.37
}   \\
    CoOp& 79.44& 41.18& \multicolumn{1}{|c}{54.24
} \\
    Co-CoOp& 77.01& 56.00& \multicolumn{1}{|c}{64.85
} \\  
    MaPLe& \textbf{80.36}& \underline{59.18}& \multicolumn{1}{|c}{\textbf{68.16}} \\   
    \midrule 
    \rowcolor{mycolor}
    APLe& \underline{79.98}& 57.93& \multicolumn{1}{|c}{\underline{67.19}} \\  
    \bottomrule
\end{tabular}
\end{minipage}
\hspace{0.1in}
\begin{minipage}[t]{0.3\textwidth}
\makeatletter\def\@captype{table}
\captionsetup{skip=2pt}
\caption*{(k) EuroSAT}
\centering
\begin{tabular}{lccc}
    \toprule
    Name     & Base& Novel & HM\\
    \midrule
    CLIP& 56.48& 64.05& \multicolumn{1}{|c}{60.03
}   \\
    CoOp& 92.19& 54.74& \multicolumn{1}{|c}{68.69
} \\
    Co-CoOp& 87.49& 60.04& \multicolumn{1}{|c}{71.21
} \\  
    MaPLe& \textbf{94.07}& \underline{73.23}& \multicolumn{1}{|c}{\underline{82.35}} \\   
    \midrule
    \rowcolor{mycolor}
    APLe& \underline{93.62}& \textbf{74.04}& \multicolumn{1}{|c}{\textbf{82.69}} \\  
    \bottomrule
\end{tabular}
\end{minipage}
\hspace{0.1in}
\begin{minipage}[t]{0.3\textwidth}
\makeatletter\def\@captype{table}
\captionsetup{skip=2pt}
\caption*{(l) UCF101}
\centering
\begin{tabular}{lccc}
    \toprule
    Name     & Base& Novel & HM\\
    \midrule
    CLIP& 70.53& 77.50& \multicolumn{1}{|c}{73.85
}   \\
    CoOp& \textbf{84.69}& 56.05& \multicolumn{1}{|c}{67.46
} \\
    Co-CoOp& 82.33& 73.45& \multicolumn{1}{|c}{77.64
} \\  
    MaPLe& 83.00& \textbf{78.66}& \multicolumn{1}{|c}{\textbf{80.77}} \\   
    \midrule 
    \rowcolor{mycolor}
    APLe& \underline{83.18}& \underline{78.13}& \multicolumn{1}{|c}{\underline{80.58}} \\  
    \bottomrule
\end{tabular}
\end{minipage}
\newline
\captionsetup{labelsep=period,skip=2pt}
\captionsetup{justification=raggedright,singlelinecheck=false}
\captionof{table}{Comparison with existing approaches. APLe demonstrates comparable generalization ability over 11 datasets.}\label{bigtable}
\vspace{-0.25cm} 
\end{table*}
\subsection{Base-to-Novel Generalization}
Table.\ref{bigtable} indicates the capability of APLe, highlighted in grey, with original CLIP, CoOp, Co-CoOp, and MaPLe over 11 datasets. We compare APLe with Co-CoOp and MaPLe separately in performance to address the diverse features of APLe.\\
\textbf{Base Classes in Generalization. }While performing prompt learning, Co-CoOp drops the average accuracy of 11 datasets in base classes from 82.69\% to 80.47\% compared to CoOp. By the independent and sequential framework, APLe stabilized the average accuracy at 81.99\% with an advanced generalization capability in all datasets compared to Co-CoOp, which demonstrates the capability in line with the others. \\
\textbf{Novel Classes in Generalization. } Compared to Co-CoOp in novel classes, owing to applying conditioning effort on the image branch and incorporating zero-shot knowledge, APLe shows significant improvement except for the OxfordPets with the 0.13\% drop in the novel class. Over 11 datasets, the average accuracy of APLe in novel classes generalization is boosted from 71.69\% to 75.11\%. Meanwhile, APLe demonstrates a 2.57\% significant increase if both base classes and novel classes are considered which is denoted as HM.
\begin{table}[H]
\vspace{-0.2cm} 
\centering
\scalebox{0.86}{%
\centering
\begin{tabular}{lcccccc}
    \toprule
    &\multicolumn{3}{c}{DTD}	& \multicolumn{3}{|c}{FGVCAircraft}		\\
       \cmidrule(lr){2-4} \cmidrule(lr){5-7}
       Sigma  & Base& Novel & HM &  \multicolumn{1}{|c}{Base}& Novel& HM\\

    \midrule
        0.02	&80.25 &	\textbf{60.55}&	\textbf{69.02} &	\multicolumn{1}{|c}{37.59}	&33.95	&35.68\\
        \rowcolor{mycolor} %
        0.05	&79.98 &	57.93&	67.19 &	\multicolumn{1}{|c}{36.65}	&34.45	&35.52\\
        0.08	&79.59 &	57.13&	66.51 &	\multicolumn{1}{|c}{37.13}	&34.45	&35.74\\
        0.10		&78.86 &	57.33&	66.39 &	\multicolumn{1}{|c}{37.17}	&35.25	&36.19\\
        0.12	&79.71 &	58.77&	67.66 &	\multicolumn{1}{|c}{\textbf{37.73}}	&34.93	&36.28\\
               \midrule
       MaPLe	&\textbf{80.36} &59.18 &68.16 &	\multicolumn{1}{|c}{37.44}	& \textbf{35.61}& \textbf{36.50}\\
         \bottomrule
\end{tabular}
}
\captionsetup{labelsep=period,skip=2pt}
\captionsetup{justification=raggedright,singlelinecheck=false}
\captionof{table}{ Effectiveness analysis of sigma $\sigma$ for two failure cases. The result highlighted denotes the values used in our case. }\label{comparison}
\vspace{-0.35cm} 
\end{table}
\noindent\textbf{APLe $vs$ MaPLe.} In Table.~\ref{bigtable} (a), it indicates that the average accuracy is comparable between APLe and MaPLe, while both approaches apply multi-modal prompt learning. APLe demonstrates the advantage in 5/11 datasets and has equivalent performance in the other 4 datasets (average gap = $\sim$0.32\%) in the novel classes. The disadvantage occurs at DTD and FGVCAircraft with 1.15\% and 1.16\% gap respectively. In general, APLe shows competitive performance in 4/11 datasets and comparable results in the other 5 datasets (average gap = $\sim$0.20\%). In DTD and FGVCAircraft, the disadvantages amounted to 0.97\% and 0.98\% respectively.
\begin{table*}[htbp]
\setlength{\tabcolsep}{3pt}
\centering
\scalebox{1}{%

\begin{tabular}{lcccccccccccc}
    \toprule
    &Source  & \multicolumn{10}{c}{Target}\\
    \cmidrule(lr){2-2} \cmidrule(lr){3-12}
     &  {ImageNet}&  {Caltech}&  {Pets}&  {Cars}&  {Flowers}&  {Food}& {Aircraft} &  {SUN}&  {DTD}&  {EuroSAT}& {UCF} & {Avg.}\\
     \midrule
    CoOp& \textbf{71.51}& 93.70& 89.14& 64.51& 68.71& 85.30& 18.47& 64.15& 41.92& 46.39 & 66.55&63.88
\\
    Co-CoOp& \underline{71.02}& \textbf{94.43}& 90.14& 65.32& \underline{71.88}& \underline{86.06}& 22.94& \textbf{67.36}& 45.73& 45.37&68.21&65.74
\\  
    MaPLe& 70.72& 93.53& \textbf{90.49}& \underline{65.57}& \textbf{72.23}& \textbf{86.20}& \textbf{24.74}& 67.01& \textbf{46.49}& \underline{48.06}& \underline{68.69}& \textbf{66.30}\\
    \midrule
    \rowcolor{mycolor}
    APLe&70.55&	\underline{93.71}&	\underline{90.44}&	\textbf{65.98}&	71.55&	86.03&	\underline{23.54}&	\underline{67.29}&	\underline{46.14}&	\textbf{48.85}&	\textbf{68.72} & \underline{66.23}\\

    \bottomrule
\end{tabular}
}

\captionsetup{labelsep=period,skip=2pt}
\captionof{table}{Comparison on cross-datasets. APLe shows a competitive result among the methods by the model trained in ImageNet.}\label{cross}
\vspace{-0.25cm} 
\end{table*}
\\
\noindent \textbf{Failure Cases.} We analyze the cases where the APLe fails while MaPLe succeeds. Specifically, Taking into account the non-learnable nature and sensitivity to image style, we assess the effectiveness of different sigma used in the Gaussian filter as shown in Table.~\ref{comparison}. The table shows that APLe enhances performance in accuracy for both base and novel classes in the DTD dataset as the sigma decreases, while the increase in sigma improves the generalization capability in FGVCAircraft. APLe achieves advanced performance when sigma is set to 0.02 for DTD and 0.12 for FGVCAircraft. Considering the nature of the Gaussian filter used in the image adapter, the results suggest that images from DTD and FGVCAircraft are situated at two distinct extremes. One necessitates preserving high-frequency details and edge features, while the other calls for more extensive smoothing to alleviate noise. Besides the analysis of sigma in image adapter, the further characteristics and advantages of APLe are explored in detail in section~\ref{Further experiments}.\noindent

\subsection{Cross-Dataset Generalization }We evaluate the generalization performance between datasets from ImageNet to other 10 datasets by training 1000 classes. Table.~\ref{cross} demonstrates the ability among CoOp, Co-CoOp and MaPLe. APLe exhibits a modest result on the source dataset and demonstrated a competitive generalization performance in 10 datasets by surpassing Co-CoOp in 6 datasets and surpassing MaPLe in 5 datasets. Ultimately, APLe outperforms 3 top records with an average accuracy of 66.23\%. This indicates that the APLe achieves comparable generalization performance with other methods in cross-dataset evaluation.
\begin{table}[H]
\centering
\scalebox{0.95}{%

\setlength{\tabcolsep}{3pt}
\begin{tabular}{lcccccc}
    \toprule
         & Source & \multicolumn{4}{c}{Target} & \\
         \cmidrule(lr){2-2} \cmidrule(lr){3-7}
          & ImageNet& -V2 & -Sketch& -A& -R  &Average\\
    \midrule
    CLIP &    66.73& 60.83&46.15& 47.77& 73.96
 &57.18
\\
    CoOp  &  \textbf{71.51}& \textbf{64.20}&47.99& 49.71& 75.21
 &59.28
\\
    Co-CoOp  &   \underline{71.02}& 64.07&48.75& \underline{50.63}&76.18
 &59.91
\\
    MaPLe &   70.72&64.07& \underline{49.15}& \textbf{50.90}&\underline{76.98} &\underline{60.28}\\
    \midrule
    \rowcolor{mycolor}
    APLe&   70.55
& \underline{64.15}&\textbf{49.24}&50.59& \textbf{77.38}&\textbf{60.34}\\    
    \bottomrule
\end{tabular}
}
\captionsetup{labelsep=period,skip=2pt}
\captionsetup{justification=raggedright,singlelinecheck=false}
\captionof{table}{Comparison in domain generalization.}\label{imagenet-a}
\vspace{-0.15cm}
\end{table}

\subsection{Domain Generlization} By training the model on the ImageNet and evaluating the modal in the out-of-domain datasets, Our approach excels in the generalization performance on out-of-distribution datasets among CoOp, Co-CoOp and MaPLe. The average results of APLe are among the top as indicated in Table.~\ref{imagenet-a}. Owing to deploying the token-wise knowledge training with adaptation for multi-modal prompts, APLe attains a competitive generalization capability among the methods.
 
\subsection{Vital Experiments and In-Depth Analysis}
\label{Further experiments}
In the aforementioned foundational experiments, APLe achieves comparable performance with the current state-of-the-art model. Another significance of APLe is that it promotes synergy for collaboration between language text and vision prompts learning to address the challenge in prompt learning due to its settings. In the following vital experiments and in-depth analysis, APLe stands out with an excellent robust and comprehensive generalization capability to adopt the V-L model compared to the baselines in downstream tasks.
\begin{table}[H]
\centering
\setlength{\tabcolsep}{3pt}
\makeatletter\def\@captype{table}
\scalebox{0.8}{%
\begin{tabular}{lccccccccc}
\toprule
		&\multicolumn{3}{c}{Both} 		&		\multicolumn{3}{|c}{Language Prompt Only}	&	\multicolumn{3}{|c}{Vision Prompt Only}\\
		\cmidrule(lr){2-4} \cmidrule(lr){5-7} \cmidrule(lr){8-10}
	&Base&	Novel	&HM	&	\multicolumn{1}{|c}{Base}	&Novel	&HM	&	\multicolumn{1}{|c}{Base}		&Novel	&HM\\
	\midrule
3	&82.33&	74.20&	78.06&	\multicolumn{1}{|c}{81.95}&	73.00&	\cellcolor{hiscolor}(77.22)&	\multicolumn{1}{|c}{82.10}&	74.63&	78.18\\
4	&81.60&	72.21&	76.62&	\multicolumn{1}{|c}{81.24}&	71.71&	\cellcolor{hiscolor}(76.18)&	\multicolumn{1}{|c}{82.04}&	74.70&	78.20\\
5	&82.27&	73.45&	77.61&	\multicolumn{1}{|c}{81.72}&	73.01&	77.12&	\multicolumn{1}{|c}{80.44}&	72.37&	\cellcolor{hiscolor}(76.19)\\
6	&82.05&	71.86&	76.62&	\multicolumn{1}{|c}{81.16}&	70.10&	\cellcolor{hiscolor}(75.23)&	\multicolumn{1}{|c}{80.40}&	71.74&	75.83\\
8	&81.94&	71.03&	76.10&	\multicolumn{1}{|c}{80.75}&	70.21&	\cellcolor{hiscolor}(75.11)&	\multicolumn{1}{|c}{81.30}&	71.67&	76.18\\
10	&82.62&	71.76&	76.81&	\multicolumn{1}{|c}{79.53}&	70.71&	\cellcolor{hiscolor}(74.86)&	\multicolumn{1}{|c}{81.83}&	71.74&	76.45\\
\midrule
AVG	&82.14&	72.42&	76.97&	\multicolumn{1}{|c}{81.06}&	71.46&	\cellcolor{hiscolor}(75.96)&	\multicolumn{1}{|c}{81.35}&	72.81&	76.84\\

\bottomrule
    \end{tabular}
    }
\captionsetup{labelsep=period,skip=2pt}
\captionsetup{justification=raggedright,singlelinecheck=false}
\captionof{table}{Prompt length analysis separated in vision and language.}\label{decople}
\vspace{-0.25cm}
\end{table}
\noindent \textbf{Prompt Length and Overfitting. }We initially explored the performance of vision and language as well as both prompts across different prompt lengths shown in Table.~\ref{decople} with a simple structure without token-wise training and adaptation.  
As the prompt length increases for both prompts or solely for the vision prompt, the generalization performances experience minor fluctuations along with a modest decrease. Meanwhile, the increase in text prompt length with leads to a noticeable decrease in accuracy. In other words, In the language prompt, extending the prompt length has a more pronounced effect, with nearly achieving the lowest scores highlighted in blue at each length. The effects are often characterized by a decrease in accuracy. The impact of decreasing performance is commonly described as overfitting in existing papers due to prompt conditions such as length. The results suggest that prompt learning behavior varies across the two different modalities, including generalization capability and the risk of knowledge conflicts. Owing to the designed framework of sequential token-wise knowledge training and token adaptation, APLe effectively improves the generalization performance and mitigates the adverse influences of challenges in the divergence of modalities token and the overfitting introduced by an increasing prompt length to adopt the V-L model.

\begin{figure}[t]
\vspace{-0.15cm}
\centering
    \includegraphics[width=1\linewidth]{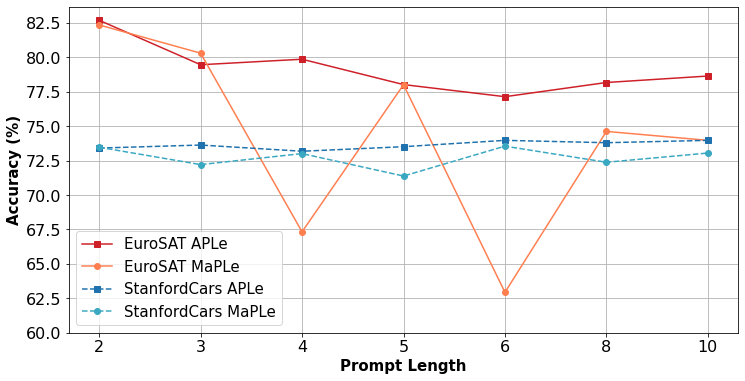} 
\captionsetup{labelsep=period,skip=2pt}
\captionof{figure}{Generalization performance comparison (HM) across various prompt lengths in the datasets, EuroSAT and StanfordCars.}\label{token-wiselengthbot}
\vspace{-0.2cm}
\end{figure}

We evaluate the generalization capability of APLe with MaPLe over different prompt lengths by two evaluation metrics. The harmonic mean indicates the generalization performance as foundational experiments, while the standard deviation effectively measures the variability in predictive results across 7 different prompt lengths. The evaluations provide an assessment of overfitting for the models. In Table.~\ref{token-wiselength}, the average performance of APLe provides significant advantages over MaPLe in both accuracy and standard deviation. The average accuracy over 7 different lengths and its standard deviation to the population are 77.07\% and 1.37 for APLe, 76.68\% and 1.92 for MaPLe. Specifically in Fig.~\ref{token-wiselengthbot}, APLe exhibits notably stable curves for the HM in EuroSAT and StanfordCars. One contains numerous textures and small objects, and the other comprises the most categories with fine-grained. 

\begin{figure}[bp]
\vspace{-0.3cm}
  \centering
    \includegraphics[width=1\linewidth]{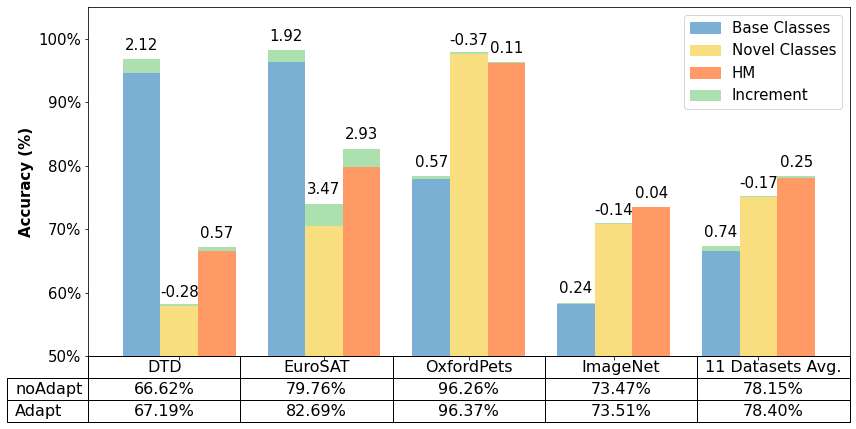} 
\captionsetup{labelsep=period,skip=2pt}
\captionsetup{justification=raggedright,singlelinecheck=false}
\captionof{figure}{Adaptation $vs$ Non-Adaptation Comparison.}\label{adapatationeffect}
\end{figure}

In the exploration of the overfitting experiment, the results resoundingly attest to the effectiveness of the framework in empowering APLe to enhance its generalization capability, showcasing outstanding robustness compared to the baseline.\\
\textbf{Effectiveness of Multi-Modal Token Adaptation. }The designed framework incorporates CLIP zero-shot knowledge to encourage APLe to learn the knowledge comprehensively and independently while alleviating overfitting challenges caused by prompt settings. However, this potentially leads to difficulties in effective collaboration between two modalities. Thus, we introduce token adaptation to promote the synergy between prompts and further enhance the generalization performance. In Fig.~\ref{adapatationeffect}, the results of the analysis in chosen datasets, DTD, EuroSAT, OxfordPets, and ImageNet, as well as average performance over 11 datasets, exhibit significant improvement in base classes and overall performance, which suggests the token adaptation supports APLe in improving the generalization capability during prompt learning and provide advanced generalization performance in downstream tasks.
\begin{table}[t]
\centering
\begin{tabular}{lrrrrr}
    \toprule
	Name&	Evl. &Base& Novel & \multicolumn{1}{|r}{HM}		\\
 \midrule
\multirow{2}{*}{\footnotesize MaPLe}&ACC.&81.88	&72.10	&\multicolumn{1}{|r}{76.68}\\
	&Std.&0.78	&2.47	&\multicolumn{1}{|r}{1.92}\\
 \midrule
 \rowcolor{mycolor}
&ACC. 	&81.86	&\textbf{72.82}	&\multicolumn{1}{|r}{\textbf{77.07}}\\
 \rowcolor{mycolor}
 \multirow{-2}{*}{ \footnotesize APLe}	&Std.&\cellcolor{mycolor}\textbf{0.64}	&\cellcolor{mycolor}\textbf{1.80}	&\multicolumn{1}{|r}{\textbf{1.37}}\\
    \bottomrule
    \end{tabular}
    \captionsetup{labelsep=period,skip=2pt}
\captionsetup{skip=2pt}
\captionof{table}{Generalization performance comparison between APLe and MaPLe across various prompt lengths in the average of 11 datasets.}\label{token-wiselength}
\vspace{-0.12cm}
\end{table}
\begin{table}[t]
\makeatletter\def\@captype{table}
\captionsetup{skip=2pt}
\caption*{\textbf{DTD}}
\centering
\resizebox{0.45\textwidth}{!}{%

\begin{tabular}{ccccccccc}
	    \toprule
\diagbox{$\lambda_d$}{$\lambda_g$}&0.3		&0.4		&0.5		&0.6		&0.7		&0.8    &\multicolumn{1}{|c}{AVG}\\
    \midrule
        0.5	&67.19	&65.79		&	63.61	&67.95	&66.85	&68.65	        &\multicolumn{1}{|c}{66.67}
        \\
        0.6	&65.21	&67.38		&	67.37	&68.13	&68.54	&68.11	        &\multicolumn{1}{|c}{67.46}\\
        0.7	&67.88	&66.92		&	67.59	&67.41	&68.13	&69.19	        &\multicolumn{1}{|c}{67.85}\\
        \rowcolor{yourcolor}
        0.8	&67.98	&68.51		&	68.15	&69.36	&69.42	&66.73	        &\multicolumn{1}{|c}{68.36}\\
    \midrule
        AVG	&67.07	&67.15		&	66.68	&68.21	&68.24	&68.17&\multicolumn{1}{|c}{67.58}	\\
    \bottomrule
    \end{tabular}
}
\vspace{+0.15cm}
\captionsetup{skip=2pt}
\caption*{\textbf{OxfordPets}}
\resizebox{0.45\textwidth}{!}{%

\begin{tabular}{cccccccc}
	    \toprule
	\diagbox{$\lambda_d$}{$\lambda_g$}&0.3		&0.4	&	0.5	&	0.6	&	0.7	&	0.8	&	\multicolumn{1}{|c}{AVG}\\
	    \midrule
 0.5	&96.37	&95.93	&96.48	&96.08	&95.79	&96.24&	\multicolumn{1}{|c}{96.15}\\
 \rowcolor{yourcolor}
0.6	&96.46	&96.49	&96.41	&96.58	&96.40	&96.25&	\multicolumn{1}{|c}{96.43}\\
0.7	&96.20	&96.32	&96.54	&96.27	&96.39	&96.24&	\multicolumn{1}{|c}{96.33}\\
0.8	&96.01	&96.40	&96.38	&96.38	&96.26	&96.65&	\multicolumn{1}{|c}{96.35}\\
    \midrule
AVG	&96.26	&96.28	&96.45	&96.33	&96.21	&96.34 &\multicolumn{1}{|c}{96.31}\\
    \bottomrule
\end{tabular}
}
\captionsetup{labelsep=period,skip=2pt}
\captionsetup{justification=raggedright,singlelinecheck=false}
\captionof{table}{Effectiveness analysis on the hyper-parameter $\lambda$ for both modalities with two chosen datasets, DTD and OxfordPets.}\label{lambdaeffect}
\vspace{-0.25cm} 
\end{table}
\newline
\noindent \textbf{Effectiveness of Hyper-Parameter $\lambda$.} We further analyze the effect of the hyper-parameter $\lambda$ ($d$ and $g$) used in integrating the result from token-wise training, language and vision respectively, with zero-shot knowledge. as described in Eq.~\ref{stage1} -~\ref{stage11}, while results are shown in Table.~\ref{lambdaeffect}. As pointed out previously, various feature representation complexities exist between the vision and language modalities, the distinctions are shown explicitly in performance and the risk of knowledge conflicts in extended experiments in prompt length. For the results shown in Table.~\ref{lambdaeffect}, the accuracy in DTD, another dataset containing textures than EuroSAT, increases with increasing values of $\lambda_d$, but the result is expected to experience fluctuations if $\lambda_g$ rise. Alternatively, the accuracy achieves higher values when $\lambda_d$ is in the vicinity of 0.8, highlighted in blue. For OxfordPets with the fewest categories with fine-grained, the result shows a similar pattern, it achieves a higher value while the $\lambda_d$ is around 0.6. The results suggest that the prompt in the language modality presents a more stable generalization capability for knowledge learning and knowledge conflict resolution than the prompt in vision.
\section{Conclusion}
The fundamental concern of prompt learning lies in the sensitivity to the setting, such as prompt lengths, while the extended prompts carry the risk of model overfitting. Likewise, the divergences between modalities contribute to the challenge of learning, which also further heightens the risk of overfitting. To this end, APLe introduced the sequential token-wise knowledge training and adaptation framework to address the challenges of multi-modal prompt learning. The proposed approach achieves comparable performance compared to the current state-of-the-art methods in three fundamental tasks. Particularly in datasets with domain shifts, the result indicates a significant gain in average accuracy. Of greater importance, the results in the analysis of the prompt length and overfitting exhibit APLe dominates the generalization performance across different prompt lengths. In light of the preceding experiments, APLe exhibits substantial generalization capability and emerges as a robust and adaptable approach for adopting the V-L model in downstream tasks.

\printbibliography

\end{document}